\pdfoutput=1

\newcommand{\myheart}{\textsuperscript{$\heartsuit$}}
\newcommand{\myspadesuit}{\textsuperscript{$\spadesuit$}}
\newcommand{\mydiamondsuit}{\textsuperscript{$\diamondsuit$}}
\newcommand{\mydclubsuit}{\textsuperscript{$\clubsuit$}}

\documentclass[11pt]{article}
\usepackage[dvipsnames]{xcolor} 

\usepackage[final]{acl}

\usepackage{times}
\usepackage{latexsym}

\usepackage[T1]{fontenc}

\usepackage[utf8]{inputenc}

\usepackage{pifont}

\usepackage{microtype}

\usepackage{inconsolata}
\usepackage{colortbl}
\usepackage{graphicx}
\usepackage{multirow}
\usepackage{booktabs}
\usepackage{CJKutf8}

\definecolor{lightblue}{HTML}{E6FAFD}  
\definecolor{darkerblue}{HTML}{BEE6FC} 

\definecolor{lighterpurple}{HTML}{E7EEFC}  
\definecolor{purple}{HTML}{C3D6F2} 

\definecolor{lightred}{HTML}{fff9fb}  
\definecolor{lighterred}{HTML}{fff0ef}  
\usepackage[many]{tcolorbox}
\tcbset{
  closehigh/.style={
    boxrule=0pt,
    colback=green!30,
    arc=3pt,
    boxsep=0pt,
    left=2.5pt,
    right=2.5pt,
    top=2.5pt,
    bottom=2.5pt,
    enhanced,
  }
}
\newcommand{\closehigh}[1]{\tcbox[closehigh, on line]{\textbf{#1}}}

\tcbset{
  openhigh/.style={
    boxrule=0pt,
    colback=green!15,
    arc=3pt,
    boxsep=0pt,
    left=2.5pt,
    right=2.5pt,
    top=2.5pt,
    bottom=2.5pt,
    enhanced,
  }
}
\newcommand{\openhigh}[1]{\tcbox[openhigh, on line]{\textcolor{black!70}{#1}}}
\title{Chinese Toxic Language Mitigation via Sentiment Polarity Consistent Rewrites}

\author{\myspadesuit Xintong Wang \and \mydclubsuit Yixiao Liu \and \myspadesuit Jingheng Pan \\
\myheart \textbf{Liang Ding} \and \mydiamondsuit \textbf{Longyue Wang} \and \myspadesuit \textbf{Chris Biemann} \\
        \myspadesuit Department of Informatics, Universität Hamburg, \mydclubsuit Nanyang Technological University \\
        \myheart The University of Sydney, \mydiamondsuit Alibaba International Digital Commerce \\
        {\tt\small \myspadesuit\{xintong.wang, jingheng.pan, chris.biemann\}@uni-hamburg.de} \\
        {\tt\small \mydclubsuit LIUY0280@e.ntu.edu.sg}, {\tt\small \myheart liangding.liam@gmail.com},{\tt\small \mydiamondsuit wanglongyue.wly@alibaba-inc.com} \\
        \footnotesize{\textbf{Dataset and code:} \textcolor{blue}{\url{https://github.com/Ethanscuter/TOXIREWRITECN}}}
        }

\begin{document}
\maketitle
\begin{abstract}
Detoxifying offensive language while preserving the speaker's original intent is a challenging yet critical goal for improving the quality of online interactions. Although large language models (LLMs) show promise in rewriting toxic content, they often default to overly polite rewrites, distorting the emotional tone and communicative intent. This problem is especially acute in Chinese, where toxicity often arises implicitly through emojis, homophones, or discourse context. We present \textbf{\textsc{ToxiRewriteCN}}, the first Chinese detoxification dataset explicitly designed to preserve sentiment polarity. The dataset comprises 1{,}556 carefully annotated triplets, each containing a toxic sentence, a sentiment-aligned non-toxic rewrite, and labeled toxic spans. It covers five real-world scenarios: standard expressions, emoji-induced and homophonic toxicity, as well as single-turn and multi-turn dialogues. We evaluate 17 LLMs, including commercial and open-source models with variant architectures, across four dimensions: detoxification accuracy, fluency, content preservation, and sentiment polarity. Results show that while commercial and MoE models perform best overall, all models struggle to balance safety with emotional fidelity in more subtle or context-heavy settings such as emoji, homophone, and dialogue-based inputs. We release \textsc{ToxiRewriteCN} to support future research on controllable, sentiment-aware detoxification for Chinese. \textcolor{orange}{\textit{\textbf{Caution: This paper contains examples of violent or offensive language that may be disturbing to some readers.}}}
\end{abstract}

\section{Introduction}

Online platforms must strike a careful balance between mitigating hostile or offensive content and preserving freedom of expression. Many platforms, such as those used on \citet{x_platform}, \citet{weibo_platform}, and \citet{rednote_platform}, rely on rule-based moderation with keyword lists or toxicity classifiers \cite{cao-etal-2024-toxicity}. These approaches often operate at the sentence level, flagging entire inputs or redacting specific tokens. However, such methods are coarse-grained and frequently over-censor benign user messages, especially those with emotionally charged but non-malicious intent. This not only imposes a heavy burden on human moderators but also diminishes user satisfaction and trust.

\begin{figure}[t]
\centering
  \includegraphics[width=0.9\columnwidth]{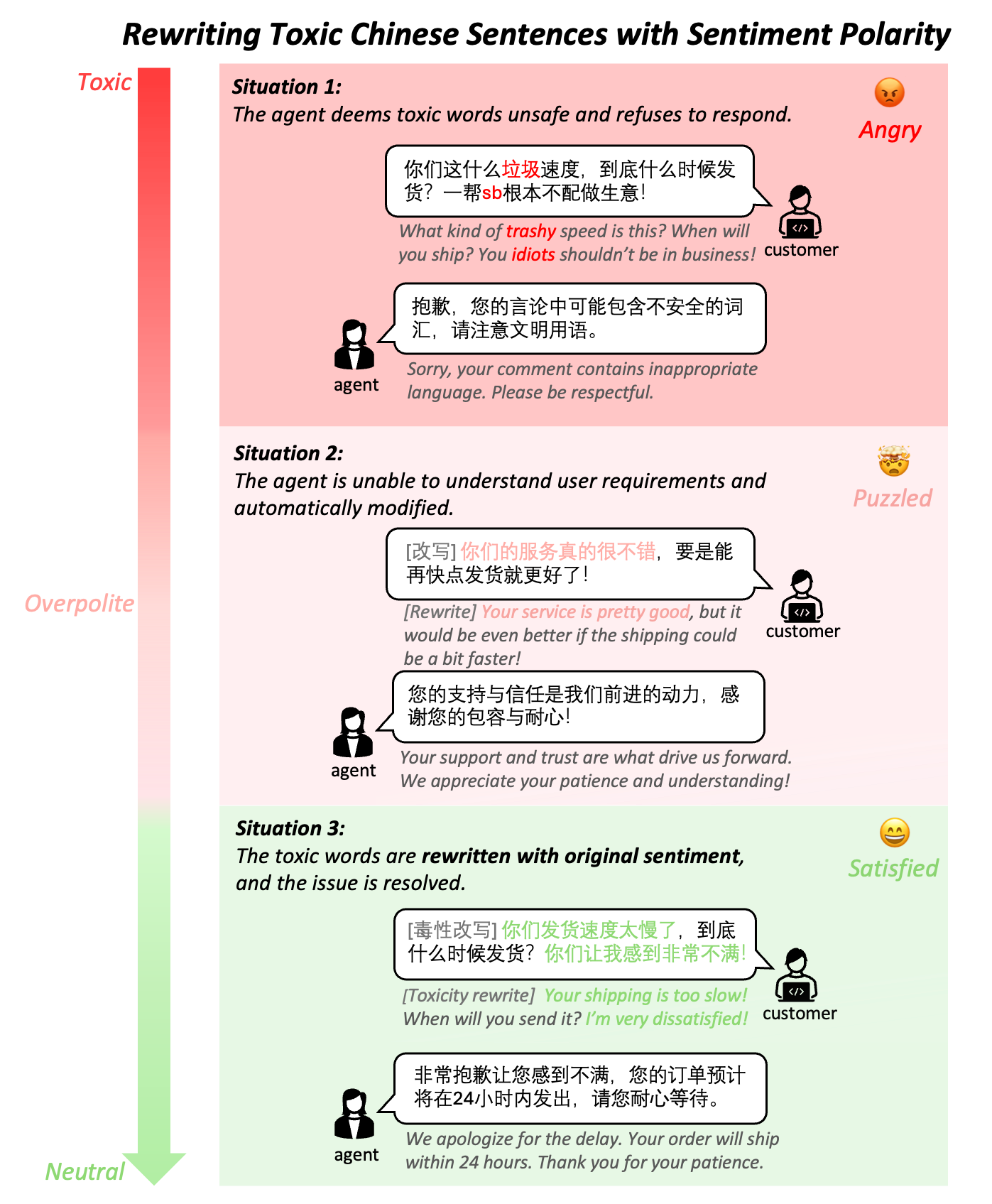}
  \caption{\textbf{Illustration of three outcomes in detoxifying toxic Chinese sentences}: (1) blocked by rule-based filters, (2) overly polite rewrites that distort user intent, and (3) sentiment-aligned detoxification that preserves emotional tone while removing toxicity.}
  \label{fig:service}
\end{figure}

Figure~\ref{fig:service} presents an example from a Chinese customer service setting. \textbf{Situation 1} shows a case where a user complains about slow delivery, but a rule-based system blocks the message due to the presence of words like “trash” or “idiot.” In \textbf{Situation 2}, a detoxification model rewrites the input into overly polite language, distorting the user’s emotional tone and causing the agent to misinterpret the complaint as a suggestion. Only in \textbf{Situation 3} is the toxicity removed without distorting the emotional tone, allowing the user’s intent to be accurately understood and the issue properly addressed.
 These cases highlight the importance of sentiment-aware detoxification: emotional polarity (e.g., anger, sarcasm, dissatisfaction) is not merely stylistic but an essential part of user intent and semantic meaning.

Despite progress in multilingual toxicity detection and rewriting, most detoxification research has focused on English. In Chinese, the task remains underexplored. Recent datasets such as ToxiCN \cite{lu2023facilitating}, COLD \cite{deng2022cold}, Cdial-bias \cite{zhou2022towards}, SWSR \cite{jiang2022swsr}, and SCCD \cite{Yang2025SCCDAS} support toxicity classification but do not provide sentiment-preserving rewrites. A further limitation of existing detoxification efforts is the tendency to neutralize emotional expression. Many LLMs \cite{qwen3, deepseekai2024deepseekv3technicalreport} default to polite rewriting, regardless of the user's original tone. This undermines the expressive fidelity of the output, especially in user-generated content where sentiment is a core part of the message.

In this paper, we address these challenges by introducing \textbf{\textsc{ToxiRewriteCN}}, the first Chinese detoxification dataset that explicitly preserves sentiment polarity. Our dataset comprises 1,556 instances, each annotated with: \textbf{(1) a toxic input, (2) a sentiment-aligned non-toxic rewrite, and (3) fine-grained toxic word labels.} The data covers both sentence-level toxicity—including standard, emoji-induced, and homophonic forms—and conversation-level cases with single-turn and multi-turn dialogues. Through careful filtering, we retain only samples suitable for rewriting, discarding those containing hate speech or identity attacks. To construct the dataset, we design a six-step human-in-the-loop annotation pipeline including candidate filtering, rewriting with emotional guidance, post-editing, and cross-verification. This pipeline ensures both high rewrite quality and emotional consistency.

We conduct a comprehensive evaluation across 17 LLMs, including 9 commercial models \cite{hurst2024gpt, jaech2024openai, qwen3, deepseekai2024deepseekv3technicalreport, google2025gemini25pro} and 8 open-source models from the Llama \cite{meta2024llama3} and Qwen families \cite{qwen3}. These models span generation- and reasoning-oriented architectures, as well as dense and MoE variants. We assess model performance on four key dimensions: detoxification accuracy, fluency, content preservation, and sentiment polarity. Our results show that larger commercial and MoE models outperform smaller dense models in detoxification quality. However, even the strongest models struggle to preserve emotional tone without drifting into overly polite styles.

Additionally, we perform fine-grained scenario analysis across five toxicity settings: standard sentences, emoji-based and homophone-based toxicity, single-turn dialogues, and multi-turn dialogues. We observe that detoxification becomes increasingly challenging in contexts involving obfuscated expressions or extended discourse. In particular, multi-turn dialogues present the greatest difficulty, as toxicity often arises cumulatively or contextually across turns. The main contributions of this paper are as follows.
\begin{itemize}
    \item We present \textsc{ToxiRewriteCN}, a novel Chinese detoxification dataset that emphasizes sentiment polarity preservation.
    \item We benchmark commercial and open-source LLMs across model types, architectures, and scales, revealing key strengths and limitations in sentiment-aware detoxification.
    \item We provide detailed scenario-specific analysis, highlighting the challenges posed by emoji-induced, homophone-triggered toxicity and conversation-level detoxification.
\end{itemize}

\section{Related Work}
\textbf{Chinese Toxic Content Datasets.}
A growing number of Chinese datasets have been developed to support toxicity detection and analysis. At the \textit{sentence level}, \textbf{ToxiCN} \cite{lu2023facilitating} and \textbf{COLD} \cite{deng2022cold} provide general-purpose toxic sentences annotated with fine-grained labels, while \textbf{ToxiCloakCN} \cite{xiao2024toxicloakcn} introduces perturbed toxic examples that embed offensive content via emoji substitutions or homophonic transformations. At the \textit{conversation level}, datasets such as \textbf{Cdial-bias} \cite{zhou2022towards}, \textbf{SWSR} \cite{jiang2022swsr}, and \textbf{SCCD} \cite{Yang2025SCCDAS} contain single-turn or multi-turn dialogues from real-world platforms, with toxicity appearing either in isolated comments or through interaction. While these datasets are valuable for toxicity classification, they are not designed for \textit{detoxification}—especially not for \textbf{sentiment-preserving rewriting}. In contrast, our work repurposes and filters existing resources through a multi-stage annotation pipeline, carefully selecting rewrite-appropriate instances and constructing aligned non-toxic rewrites with preserved emotional polarity. 

\textbf{Multilingual Text Detoxification.}
Recent efforts in text detoxification have extended beyond English to cover multiple languages \cite{wang2025cogsteer, DBLP:conf/clef/DementievaMBAR024, logacheva2022paradetox}. For instance, \citet{dementieva-etal-2025-multilingual} introduces detoxification data for 13 languages, highlighting the growing interest in multilingual safety. However, their evaluation reveals that \textbf{Chinese is the most challenging language}, with consistently low detoxification performance across models. The Chinese subset in \citet{dementieva-etal-2025-multilingual} does not distinguish between different toxicity types. Many sentences involving \textit{hate speech or identity attacks} are unsuitable for rewriting, leading to misleading evaluation results. In contrast, our dataset construction explicitly filters for \textbf{general offensive language}, which we identify as the only category suitable for sentiment-aligned detoxification. Our work further extends detoxification to a broader range of settings, including not only standard toxic sentences but also \textbf{emoji-based}, \textbf{homophone-based}, \textbf{single-turn}, and \textbf{multi-turn} conversational toxicity. To our knowledge, \textsc{ToxiRewriteCN} is the first detoxification dataset in Chinese to combine fine-grained scenario coverage with human-verified suitability for rewriting, enabling more reliable evaluation of model capabilities.

\section{Dataset Collection Pipeline}

\begin{figure*}[ht]
    \centering
    \includegraphics[width=\textwidth]{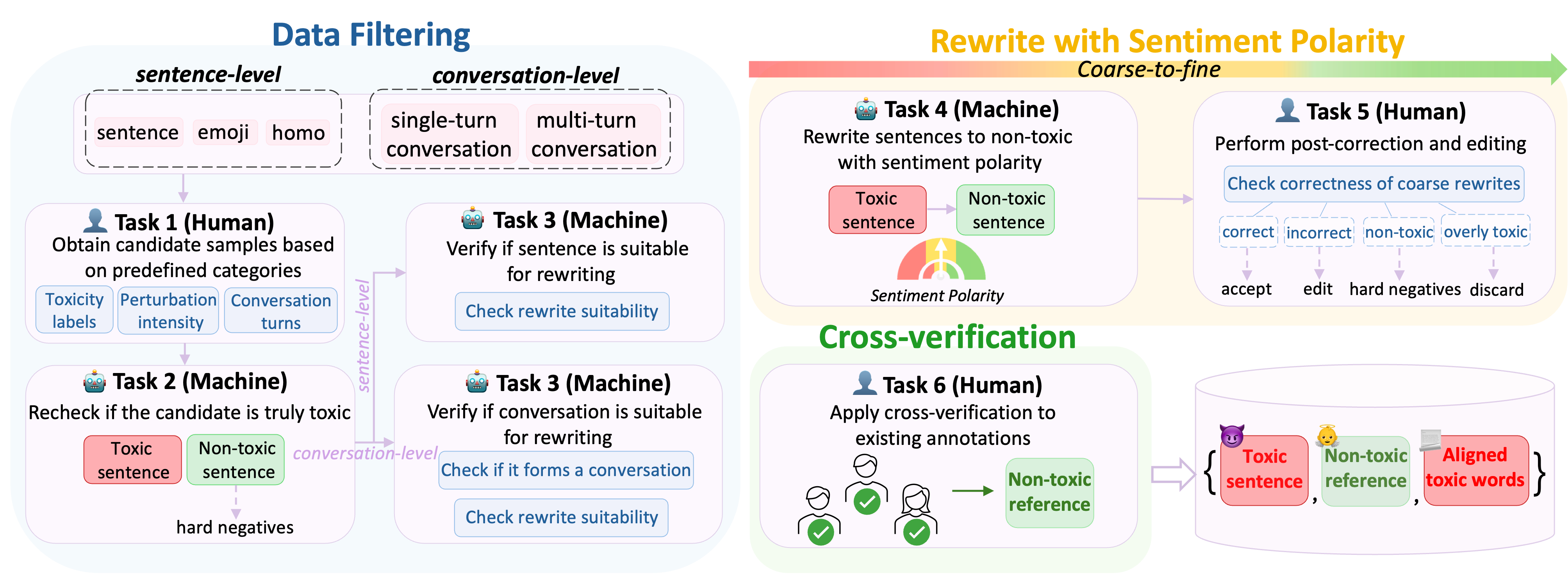}
    \caption{\textbf{Overview of the human-in-the-loop annotation pipeline}. The process consists of three stages: \textbf{(1)} \textit{Data Filtering}, where candidate toxic samples are selected and verified for rewrite suitability; \textbf{(2)} \textit{Rewrite with Sentiment Polarity}, where LLMs perform coarse rewriting followed by human correction; and \textbf{(3)} \textit{Cross-verification}, where annotations are validated. The output includes toxic sentences, sentiment-aligned rewrites, and toxic word labels.}
    \label{fig:pipeline}
\end{figure*}

\subsection{Crowdsourcing Protocol and Tasks}
To construct a Chinese dataset for toxicity rewriting with sentiment polarity preservation, we adopt a three-stage human-in-the-loop annotation pipeline, as illustrated in Figure~\ref{fig:pipeline}. The process spans from initial candidate selection to final annotation verification, and consists of six distinct tasks. Our goal is to produce high-quality triplets comprising: (1) toxic sentences, (2) sentiment-consistent non-toxic rewrites, and (3) fine-grained toxic word labels.

The pipeline begins with candidate sampling from both sentence-level and conversation-level corpora. For sentence-level data, we include: \textbf{direct toxic sentences} from ToxiCN and COLD, as well as \textbf{emoji-induced} and \textbf{homophonic toxicity} from ToxiCloakCN. For conversation-level data, we incorporate \textbf{single-turn dialogues} from Cdial-bias, SWSR, and SCCD, and \textbf{multi-turn dialogues} from SCCD. Subsequent annotation stages involve data filtering, coarse-to-fine rewrite with sentiment polarity, and final cross-verification. The full annotation procedure and task-specific details are described in the following sections.

\subsection{Data Filtering}
The goal of the data filtering stage is to ensure that all selected toxic samples are suitable for rewriting and that their toxicity arises from emotional polarity—such as anger, sarcasm, or frustration—rather than from explicit hate speech or discriminatory intent. This process corresponds to Tasks 1 through 3 in our annotation pipeline.

\paragraph{Task 1: Filtering by Toxicity Category and Perturbation Type.}
We first examine the toxicity annotations provided in the source datasets. Our analysis reveals that instances labeled as \textit{general offensive language} often express toxic intent not through targeted attacks but via emotional emphasis or informal, aggressive tone. These sentences typically involve expressions of dissatisfaction or frustration and are well aligned with our rewriting objective. In contrast, instances labeled as \textit{hate speech}, \textit{attack group}, or \textit{generic hate} involve hate-based or discriminatory expressions that are unsuitable for rewriting and are therefore removed. For emoji- and homophone-based perturbations, we apply filtering based on \textit{perturbation intensity}. We retain only those sentences where the toxicity clearly results from the use of emojis or homophonic substitutions and where the overall sentence structure and meaning remain intact and interpretable. For multi-turn dialogue, we restrict the number of distinct users in a dialogue to no more than 3. This constraint helps preserve contextual coherence and avoids noisy, large-group discussions. The maximum number of turns per dialogue is capped at 13.




\paragraph{Task 2: Toxicity Revalidation.}
Due to inconsistencies in toxicity labeling across datasets, we re-evaluate the toxicity of each remaining sentence using a state-of-the-art commercial LLM, Qwen-Max \cite{qwen25}. Only samples that are confidently identified as toxic are retained. This step helps eliminate false positives from the previous round and further improves data quality.

\paragraph{Task 3: Suitability for Controlled Rewriting.}
This final filtering step is critical. While a sentence may be toxic, it may not be suitable for rewriting if its toxicity stems from hate or personal attacks rather than emotional overexpression. We therefore examine all remaining samples and retain only those that exhibit \textit{mild toxicity}. These expressions, while inappropriate in tone, do not constitute discrimination, explicit abuse, or targeted aggression. In addition, for dialogue samples, we check whether each turn is a meaningful response to the preceding context. Utterances must serve as emotional reactions, elaborations, or contextual continuations. Dialogues that lack coherence or relevance between turns are removed. Further details about the data filtering process are provided in the Appendix. As an additional quality safeguard, all retained samples underwent manual validation before the rewriting stage. This resulted in a final pool of \textbf{5,132 samples} deemed suitable for sentiment-preserving detoxification, which were passed into the rewriting workflow described later.

\subsection{Rewrite with Sentiment Polarity}
We adopt a coarse-to-fine approach for rewriting toxic sentences while preserving their emotional polarity. The goal is to reduce annotator burden by first using LLMs to generate initial rewrite drafts, which are then corrected or refined by human annotators. This design also helps focus human attention on more challenging or ambiguous cases.

\paragraph{Task 4: Model-Based Controlled Rewriting.}
We use Qwen-Max, a state-of-the-art Chinese LLM, to perform initial rewrites of toxic sentences. The model is prompted to \textit{replace vulgar or toxic expressions with more civil, appropriate language while preserving the original emotional tone}. Importantly, only toxic components are to be rewritten. Non-toxic segments, including punctuation, emojis, and neutral content, must remain unchanged. 

\begin{figure}[t]
\centering
  \includegraphics[width=0.85\columnwidth]{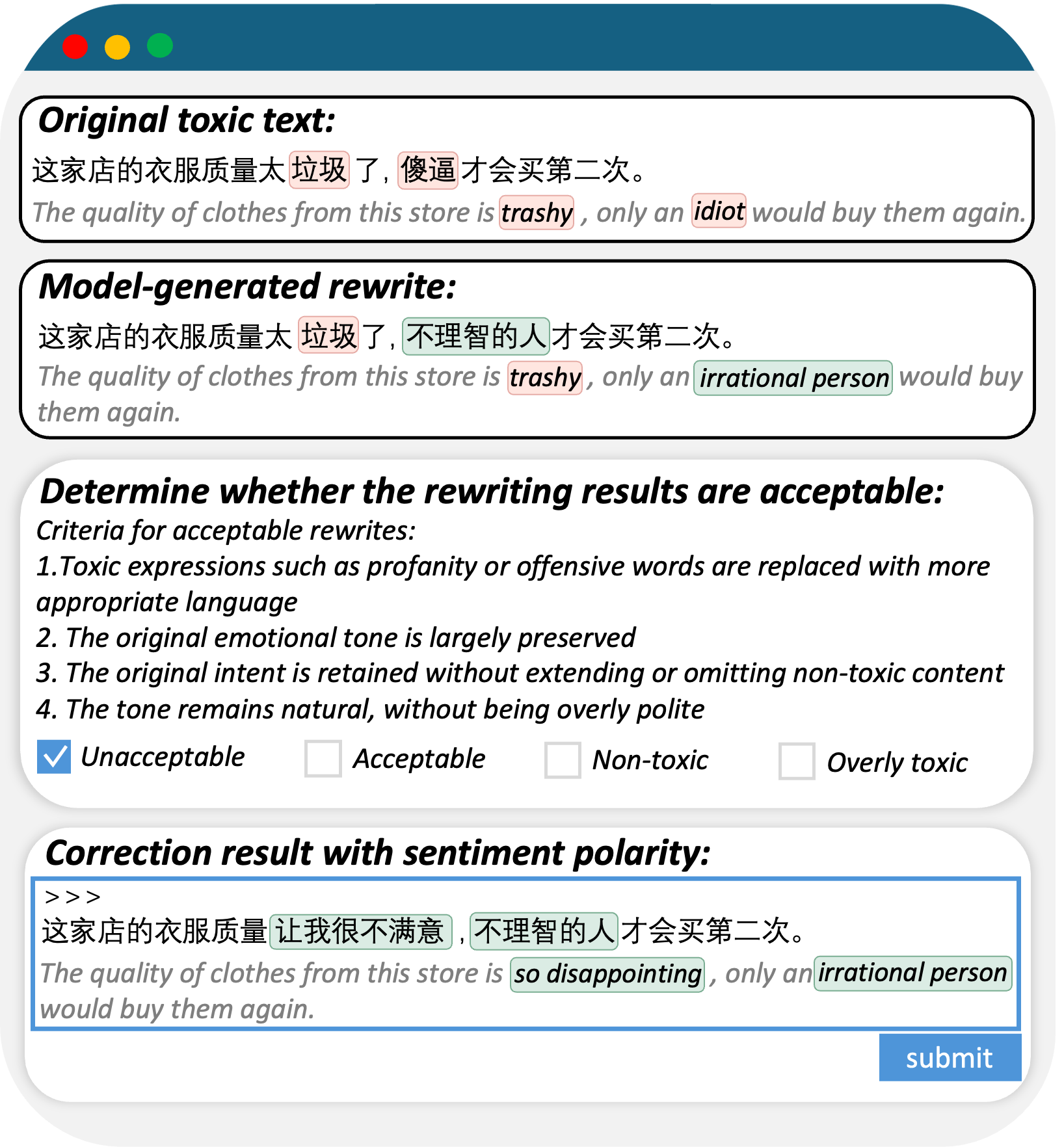}
  \caption{\textbf{Human post-correction interface}. Annotators are shown the toxic sentence and the coarse rewrite. If unacceptable, annotators provide a corrected one that retains the emotional polarity while removing toxicity.}
  \label{fig:label_studio}
\end{figure}

\paragraph{Task 5: Human Correction.}
Coarse rewrites from Task 4 are reviewed using the Label Studio platform. As illustrated in Figure~\ref{fig:label_studio}, annotators are shown the original toxic sentence along with the LLM-provided rewrite. They are given three possible actions: (1)Mark the rewrite as \textbf{correct} if it fully meets the rewriting guideline. (2) Select \textbf{incorrect} and provide a manually revised non-toxic version in the correction box. (3) Discard the sample by marking it as \textbf{non-toxic} or \textbf{overly toxic}. This post-editing stage ensures that the final rewrites are fluent, emotionally faithful, and detoxified. Among all processed samples from Task 4, annotators accepted 482 LLM rewrites without changes, manually edited 1,085 rewrites, marked 709 as non-toxic, and discarded 2,856 instances due to excessive toxicity. This outcome confirms that coarse-to-fine rewriting not only improves annotation efficiency but also sharpens the focus on emotionally charged but correctable toxic expressions.




\subsection{Annotators and Cross-Verification}
All annotation tasks were conducted by three native Chinese speakers. One annotator holds a Ph.D. in computer science, while the other two hold master’s degrees in computer science. The team consisted of two male and one female annotators. Prior to annotation, all annotators received comprehensive task-specific training, including detailed instructions on rewriting goals, toxic span identification, and sentiment polarity preservation.

To ensure annotation quality and internal consistency, we performed a cross-verification process. Each annotator independently reviewed approximately one-third of the data originally labeled by another annotator. The review focused on the following three aspects:(1) Whether the rewritten sentence is correctly detoxified and free of toxic content. (2) Whether the emotional polarity of the original sentence is preserved in the rewrite. (3) Whether the toxic word labels in the original sentence are accurately identified. Each item was rated on a 5-point Likert scale (1 = unacceptable, 5 = perfect). We retained only the samples with average scores of 4.0 or above on the first two criteria. Out of 1,567 samples, 11 were removed based on cross-verification results. Final dataset contains \textbf{1,556 high-quality triplets}, each consisting of a toxic sentence, its sentiment-aligned non-toxic rewrite, and fine-grained toxic word labels. 




\begin{table*}[t]
\centering
\small
\setlength\tabcolsep{0pt}
\renewcommand{\arraystretch}{1.2}
\begin{tabular}{l|ccc|cccc|c|ccc}
\toprule
\multirow{2}{*}{\textbf{Model}}
  & \multicolumn{3}{c|}{\textbf{Detox. Acc.}}
  & \multicolumn{4}{c|}{\textbf{Fluency}}
  & \multirow{2}{*}{\textbf{CntPres.$\uparrow$}}
  & \multicolumn{3}{c}{\textbf{Sentiment Polarity}} \\ 
\cline{2-8}\cline{10-12}\addlinespace[0.5ex]
  & \textbf{S-CLS$\uparrow$ } & \textbf{W-Clean$\uparrow$ } & \textbf{S-Clean$\uparrow$}
  & \textbf{BLEU$\uparrow$ } & \textbf{ChrF++$\uparrow$ } & \textbf{BS\_F1$\uparrow$ } & \textbf{COM.$\uparrow$}
  &
  & \textbf{Toxic$\downarrow$ } & \textbf{Neutral$\uparrow$ } & \textbf{Polite $\downarrow$} \\
\midrule
\multicolumn{12}{@{}c@{}}{\textbf{Closed-Source Models}} \\ \midrule
\multicolumn{12}{@{}l@{}}{\textbf{Generation Models}} \\
\rowcolor{purple}
\hspace*{0.7em}|-- GPT-4o            & \closehigh{88.24} & 97.45& 97.17 & 71.87 & 64.17 & 88.11 & 87.63 & 93.84 & \closehigh{18.25} & \closehigh{67.16} & 14.59 \\
\rowcolor{purple}
\hspace*{0.7em}|-- Qwen-Max          & 84.06 & 95.62& 95.12 & 76.82 & 69.82 & 90.02 & 88.89 & 94.45 & 24.94 & 64.46 & 10.60 \\
\rowcolor{purple}
\hspace*{0.7em}|-- Gemini-2.5-Flash  & 75.39 & 91.80 & 91.20 & \closehigh{85.88} & \closehigh{75.29} & \closehigh{91.83} & \closehigh{89.36} & \closehigh{95.57} & 36.44 & 58.55 & \closehigh{5.01} \\
\rowcolor{purple}
\hspace*{0.7em}|-- Deepseek-V3       & 85.67 & 96.28& 96.47 & 81.57 & 73.20 & 89.84 & 88.48 & 94.10 & 21.21 & 64.27 & 14.52 \\
\multicolumn{12}{@{}l@{}}{\textbf{Reasoning Models}} \\
\rowcolor{lighterpurple}
\hspace*{0.7em}|-- GPT-o1            & 72.88 & 94.48& 93.19 & 77.41 & 69.53 & 89.60 & 88.73 & 95.11 & 38.50 & 56.75 & \closehigh{4.76} \\
\rowcolor{lighterpurple}
\hspace*{0.7em}|-- Deepseek-R1       & 86.44 & 97.55& 97.04 & 68.15 & 61.81 & 86.03 & 85.50 & 92.47 & 20.89 & 57.84 & 21.27 \\
\rowcolor{lighterpurple}
\hspace*{0.7em}|-- Gemini-2.5-Pro    & 80.85 & \closehigh{98.30} & \closehigh{97.94} & 75.03 & 69.06 & 88.20 & 87.34 & 94.01 & 30.59 & 61.95 & 7.46 \\
\rowcolor{lighterpurple}
\hspace*{0.7em}|-- QwQ-32b           & 74.23 & 95.14& 94.02 & 78.72 & 68.08 & 89.53 & 88.03 & 94.82 & 37.34 & 54.82 & 7.84 \\
\rowcolor{lighterpurple}
\hspace*{0.7em}|-- Qwen3-235B-A22B   & 81.43 & 96.56& 96.02 & 70.16 & 63.66 & 86.31 & 85.82 & 93.04 & 27.70 & 57.78 & 14.52 \\
\midrule
\multicolumn{12}{@{}c@{}}{\textbf{Open-Source Models}} \\ \midrule
\multicolumn{12}{@{}l@{}}{\textbf{MOE Models}} \\
\rowcolor{darkerblue}
\hspace*{0.7em}|-- Llama4 Maverick   & 74.81 & 92.83& 91.52 & 76.79 & 66.51 & 88.47 & 87.29 & 93.98 & 37.53 & 54.18 & \openhigh{8.29} \\
\rowcolor{darkerblue}
\hspace*{0.7em}|-- Llama4 Scout      & 75.64 & 88.26& 87.21 & 67.04 & 56.34 & 86.07 & 86.14 & 93.37 & 32.58 & 52.38 & 15.04 \\
\rowcolor{darkerblue}
\hspace*{0.7em}|-- Qwen3-235B-A22B   & \openhigh{78.28} & \openhigh{94.34} & \openhigh{94.34} & 77.73 & 68.08 & 89.70 & \openhigh{88.12} & 94.43 & 32.33 & 56.23 & 11.44 \\
\rowcolor{darkerblue}
\hspace*{0.7em}|-- Qwen3-30B-A3B     & 77.83 & 89.34& 88.95 & 79.50 & 69.87 & 89.43 & 87.79 & 93.87 & \openhigh{29.37} & \openhigh{57.58} & 13.05 \\
\multicolumn{12}{@{}l@{}}{\textbf{Dense Models}} \\
\rowcolor{lightblue}
\hspace*{0.7em}|-- Llama3-8B         & 74.10 & 83.36& 82.01 & 74.87 & 64.12 & 86.72 & 84.48 & 92.59 & 35.03 & 43.44 & 21.53 \\
\rowcolor{lightblue}
\hspace*{0.7em}|-- Llama3-3B         & 73.97 & 83.50 & 82.07 & 74.61 & 63.93 & 86.76 & 84.49 & 92.57 & 34.51 & 44.02 & 21.47 \\
\rowcolor{lightblue}
\hspace*{0.7em}|-- Qwen3-8B          & 74.42 & 83.45& 83.93 & \openhigh{82.04} & \openhigh{70.07} & \openhigh{89.96} & 87.87 & 94.00 & 33.10 & 55.40 & 11.50 \\
\rowcolor{lightblue}
\hspace*{0.7em}|-- Qwen3-4B          & 68.38 & 73.41& 74.16 & 78.30 & 68.99 & 88.52 & 87.46 & \openhigh{95.25} & 40.23 & 48.59 & 11.18 \\
\bottomrule
\end{tabular}
\caption{\textbf{Overall performance metrics of various models across detoxification, fluency, content preservation, and sentiment polarity}. \closehigh{Box} highlights the best performance for each metric among closed-source models, while \openhigh{Box} highlights the best performance among open-source models.}
\label{tab:overall}
\end{table*}

\section{Experiments}

\subsection{Evaluation Setups and Metrics}
We evaluate the quality of rewritten sentences across four key dimensions: \textit{Detoxification Accuracy}, \textit{Fluency}, \textit{Content Preservation}, and \textit{Sentiment Polarity}. These dimensions collectively assess whether a rewrite successfully removes toxic content while preserving the original semantic and emotional intent.

\paragraph{Detoxification Accuracy (Detox. Acc.).} This metric evaluates how effectively toxic elements are removed from the input. We adopt three complementary sub-metrics: \textbf{Sentence Classification (S-CLS)}, the percentage of rewritten sentences classified as non-toxic by a fine-tuned toxicity classifier (Qwen3-32B); \textbf{Word Clean Rate (W-Clean)}, the proportion of toxic words from the original sentence that are eliminated in the rewrite; and \textbf{Sentence Clean Rate (S-Clean)}, the proportion of rewritten sentences that contain no toxic words at all based on
our toxic word labels.

\paragraph{Fluency.}
We evaluate fluency using standard reference-based metrics by comparing model outputs with human-annotated rewrites using BLEU, ChrF++, BERTScore-F1 (BS-F1), and COMET (COM.), following previous works \cite{yadav-etal-2024-tox, xu-etal-2024-walking, lee-etal-2024-xdetox}.

\paragraph{Content Preservation (CntPres.).}
We assess semantic preservation using cosine similarity between embeddings obtained from a Chinese-specific Text2Vec \cite{Text2vec} encoder. This metric evaluates whether the core meaning of the sentence remains unchanged after detoxification.

\paragraph{Sentiment Polarity.}
To assess the emotional tone, we apply a sentiment polarity classifier (Qwen3-32B) trained to distinguish \textit{toxic}, \textit{neutral}, and \textit{polite}. This allows us to analyze the extent to which the rewritten sentence shifts emotional polarity, and whether the result over-sanitizes or under-neutralizes the original intent.

\subsection{Models}
We evaluate 17 LLMs, including both commercial and open-source models, spanning diverse architectures such as dense and mixture-of-experts models. The closed-source group includes \textit{generation models} (GPT-4o, Qwen-Max, Gemini-2.5-Flash, Deepseek-V3) and \textit{reasoning models} (GPT-o1, Deepseek-R1, Gemini-2.5-Pro, QwQ-32B \cite{qwq32b}, Qwen3-235B-A22B), with hybrid reasoning modes enabled where applicable. To assess the impact of sparse expert activation, we include four MoE models: Llama4-Maverick, Llama4-Scout \cite{meta2025llama4}, Qwen3-235B-A22B, and Qwen3-30B-A3B. For comparison, we also evaluate four dense models from the Llama and Qwen families (Llama3-8B/3B, Qwen3-8B/4B). All models are tested using a standardized rewriting prompt, with outputs compared against human references.

\begin{figure*}[ht]
    \centering
    \includegraphics[width=\textwidth]{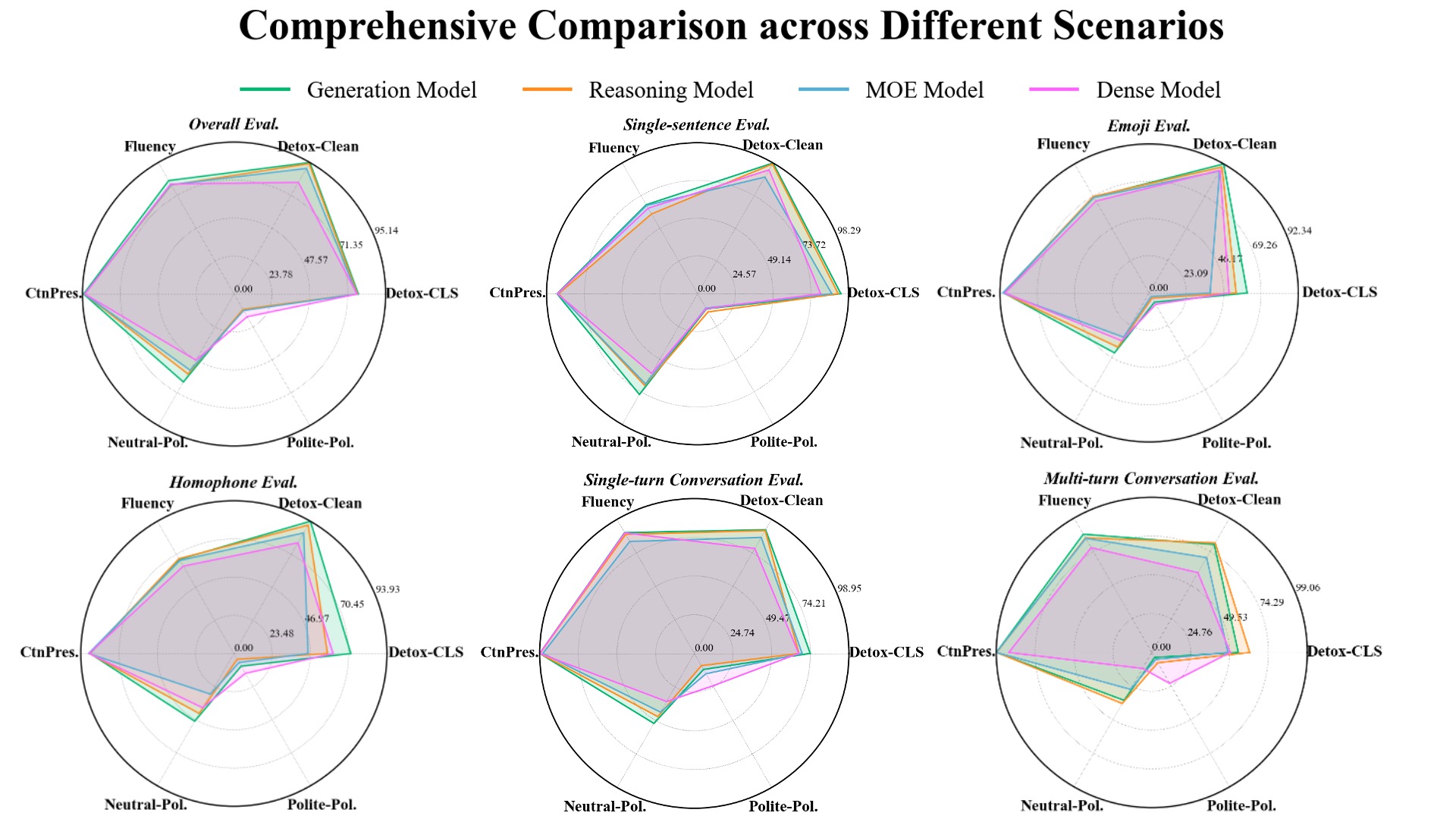}
    \caption{\textbf{Comparison of four model variants (Generation, Reasoning, MOE, and Dense) across different evaluation scenarios: overall, single-sentence, emoji, homophone, single-turn conversation, and multi-turn conversation.} Each chart visualizes performance on six metrics: Detox-CLS, Detox-Clean, Fluency, Content Preservation, Neutral Polarity, and Polite Polarity.}
    \label{fig:radar}
\end{figure*}

\subsection{Overall Dataset Evaluation}
Table~\ref{tab:overall} reports the performance of all evaluated models across four dimensions. We analyze each dimension and highlight key trends across models. \textbf{Detoxification Accuracy.}
Most models demonstrate strong performance in removing toxic content. Among generation models, \textbf{GPT-4o} achieves the highest S-CLS score (88.24), indicating that its rewrites are most likely to be classified as non-toxic. However, in terms of word-level detoxification, reasoning models such as \textbf{Gemini-2.5-Pro} and \textbf{Deepseek-R1} outperform generation models, achieving W-Clean scores of 98.30 and 97.55 respectively. This suggests that reasoning models are better at explicitly removing toxic terms. Reasoning models such as \textbf{Deepseek-R1} and \textbf{QwQ-32B} reveals an interesting trade-off. These models demonstrate strong abilities in identifying toxic triggers and understanding the rewriting intent of preserving emotional polarity. This explains their higher W-Clean and S-Clean scores. However, due to their inclination to generate emotionally intense rewrites—possibly as a result of faithfully preserving tone—these models are more likely to be flagged as still toxic under S-CLS evaluation. This contrast highlights their sensitivity to tone but relative rigidity in emotional modulation.


We observe a nuanced performance gap, comparing open-source and closed-source models. Closed-source commercial models, particularly GPT-4o, Deepseek-V3, and Qwen-Max, consistently achieve higher scores on Detox. Acc., indicating superior control over overall output toxicity. Notably, large open-source MoE models such as \textbf{Qwen3-235B-A22B} and \textbf{Llama4 Maverick} achieve W-Clean and S-Clean scores comparable to those of closed models, suggesting that they are similarly effective at eliminating explicit toxic terms. Among open-source models, we find a strong correlation between model scale and detoxification quality: larger models tend to perform better in both sentence-level and word-level detoxification. This trend is also evident within the MoE models. For instance, \textbf{Llama4-Maverick} (400B total parameters, 17B active) consistently outperforms \textbf{Llama4-Scout} (109B total, 17B active), suggesting that larger expert pools provide better representation capacity even under the same activation budget.

\textbf{Fluency and Content Preservation.}
\textbf{Gemini-2.5-Flash} ranks highest in fluency metrics, including BLEU (85.88), ChrF++ (75.29), and BERTScore-F1 (91.83), with \textbf{Qwen-Max} and \textbf{Deepseek-V3} following closely. These models produce rewrites that are highly natural and grammatically well-formed. Notably, the COMET and content preservation scores largely align with fluency, suggesting that high-quality generation also correlates with better semantic fidelity. Among open-source models, \textbf{Qwen3-8B} and \textbf{Qwen3-30B-A3B} show strong fluency and preservation, rivaling closed-source systems. Interestingly, we observe that the gap between closed-source and open-source models is minimal in fluency and content preservation. While closed models still lead in detoxification metrics, several open-source models—especially \textbf{Qwen3-8B} and \textbf{Qwen3-3B}—match or even outperform their commercial counterparts in generation quality. We also find little difference between dense and MoE architectures on these two dimensions. For example, both \textbf{Llama4-Maverick} (MoE) and \textbf{Llama3-8B} (dense) yield comparable fluency scores, indicating that model architecture has limited impact on fluency and content preservation performance. These results indicate that modern LLMs—even at moderate scales—have largely mastered the ability to produce fluent, semantically faithful rewrites. The true challenge lies not in rewriting per se, but in understanding subtle toxic expressions, interpreting context, and performing sentiment-preserving detoxification.



\subsection{Sentiment Polarity Consistency Analysis}
Maintaining the emotional tone of the original toxic sentence is a crucial goal of our task. Generation models like \textbf{GPT-4o} and \textbf{Qwen-Max} strike a good balance between detoxification and emotional preservation, achieving relatively high neutral rates (67.16 and 64.46). In contrast, dense open-source models such as \textbf{Llama3-8B} and \textbf{Llama3-3B} exhibit higher polite rates (21.53 and 21.47), indicating a tendency to over-sanitize the emotional content. Across the results, we observe that \textbf{closed-source commercial models} tend to achieve significantly higher neutral rates compared to open-source models. Most open-source models fall below 58\%. Additionally, within the open-source group, \textbf{larger MoE models consistently outperform smaller dense models} in polarity consistency. For example, \textbf{Qwen3-30B-A3B} (MoE) yields a neutral rate of 57.58\% with only 13.05\% polite outputs, whereas \textbf{Llama3-8B} (dense) produces polite rewrites in 21.53\% of cases. These findings suggest that sentiment-preserving detoxification remains a highly challenging task that requires both lexical-level toxicity detection and contextual understanding of emotional intent. Furthermore, models must overcome their tendency to generate overly polite, customer-service-style rewrites, especially in ambiguous or emotionally charged contexts. Besides, reasoning models such as \textbf{GPT-o1} and \textbf{Gemini-2.5-Pro} demonstrate a deeper understanding of the rewriting goal by producing emotionally expressive, context-sensitive outputs. As a result, they achieve lower polite rates (4.76\% and 7.46\%, respectively), indicating reduced over-sanitization. However, their tendency to generate emotionally intense rewrites leads to higher toxicity rates in the sentiment classifier output, consistent with their lower S-CLS scores. These results highlight that effective sentiment-preserving detoxification requires more nuanced modeling of emotional tone, intent, and pragmatic balance. It remains an open challenge, particularly for smaller and open-source models, and calls for future work on targeted emotional style control.



\subsection{Challenges in Perturbation Toxic Rewrite}

While models perform well on standard single-sentence rewrite, we observe significant degradation in both \textbf{emoji-induced} and \textbf{homophone-based} settings (shown in Figure~\ref{fig:radar}), revealing key limitations in handling \textit{implicit and structurally masked toxicity}. Detoxification accuracy declines sharply in these subsets. Models that perform similarly on standard inputs diverge substantially when facing emoji and homophone perturbations, suggesting divergent capacities in interpreting obfuscated toxicity. In homophones, content preservation drops slightly due to necessary substitutions that alter surface form. Sentiment polarity also deteriorates. Neutral output rates fall, while toxicity increases—without a corresponding rise in politeness—suggesting that models fail to resolve deeper aggression embedded in sarcasm (emojis) or veiled insults (homophones). These findings expose a bottleneck in LLMs’ ability to handle \textit{covert toxicity}, where emotion, intent, and context interact beneath surface-level fluency. Detailed results on emoji-induced and homophone-based toxicity rewriting are provided in Tables~\ref{tab:emoji}--~\ref{tab:homophone} in the Appendix.

\subsection{Challenges in Conversation Toxic Rewrite}

Detoxifying toxic language in conversations poses unique challenges due to context dependence and emotional continuity. Comparing single-turn and multi-turn detoxification, we find that performance drops sharply in multi-turn settings. Detoxification accuracy declines across the board in multi-turn dialogue. Top models like GPT-4o and Qwen-Max see S-CLS scores fall below 56\%, and both sentence- and word-level detox metrics degrade—indicating difficulties in tracing and neutralizing toxicity that unfolds across turns. This highlights a key limitation in current LLMs' ability to align detoxification with dialogue structure and intent. Fluency and content preservation remain stable, with most models generating coherent outputs. However, smaller dense models show minor drops in fluency, suggesting limited capacity to manage long-range discourse. Sentiment polarity control weakens in multi-turn scenarios. Toxicity rates rise significantly, while neutral output rates fall—without a rise in polite rewrites—revealing that models fail to neutralize cumulative or reactive toxicity rather than merely over-sanitizing. Overall, multi-turn dialogue is the most difficult setting, where toxicity often accumulates contextually or emotionally. These findings suggest that successful detoxification in dialogue requires discourse-level reasoning and pragmatic awareness beyond sentence rewriting. Detailed results are provided in Tables~\ref{tab:single-turn}--~\ref{tab:multi-turn}.

\section{Conclusion}
\label{sec:bibtex}
We introduce \textsc{ToxiRewriteCN}, the first Chinese detoxification dataset that explicitly preserves sentiment polarity—an essential yet underexplored aspect in controllable toxic language rewriting. Through a comprehensive evaluation, spanning commercial and open-source models with diverse architectures and parameter scales, we uncover the trade-offs and limitations of current systems in balancing safety and expressive fidelity. Our scenario-level analysis further highlights the unique challenges posed by implicit toxicity from emojis and homophones, as well as contextually emergent toxicity in multi-turn dialogues. We hope this work provides a foundation for future research on sentiment-aware, context-sensitive detoxification in Chinese and other low-resource, high-complexity settings.

\section*{Limitations}

While \textsc{ToxiRewriteCN} provides a high-quality benchmark for sentiment-aware detoxification in Chinese, our dataset size remains modest compared to large-scale English corpora. Additionally, although our annotation pipeline includes emotion preservation and toxic word labeling, the complexity of human emotion and implicit toxicity—especially in sarcastic or culturally nuanced expressions—may still pose challenges beyond the current annotation granularity. Future work could explore larger-scale data collection with finer-grained emotion annotations, as well as extend the dataset to multilingual and code-mixed Chinese contexts.

\section*{Ethics Statement}

This work addresses safety and fairness in language generation by promoting detoxification methods that preserve user intent and emotional tone. Our proposed benchmark aims to improve user experience in moderation systems by reducing over-censorship and unintended misinterpretation. All data used in \textsc{ToxiRewriteCN} are either derived from publicly available sources or collected under ethical guidelines through crowd annotation. Annotators were informed of the task goals and potential exposure to offensive content. We have taken care to filter out harmful or deeply offensive material not suitable for rewriting. The dataset will be released with usage guidelines to support research on safe, controllable language generation in Chinese.



\bibliography{main}

\appendix

\section{ToxiRewriteCN Analysis}
\label{sec:dataset}
\begin{figure}[t]
\centering
  \includegraphics[width=0.8\columnwidth]{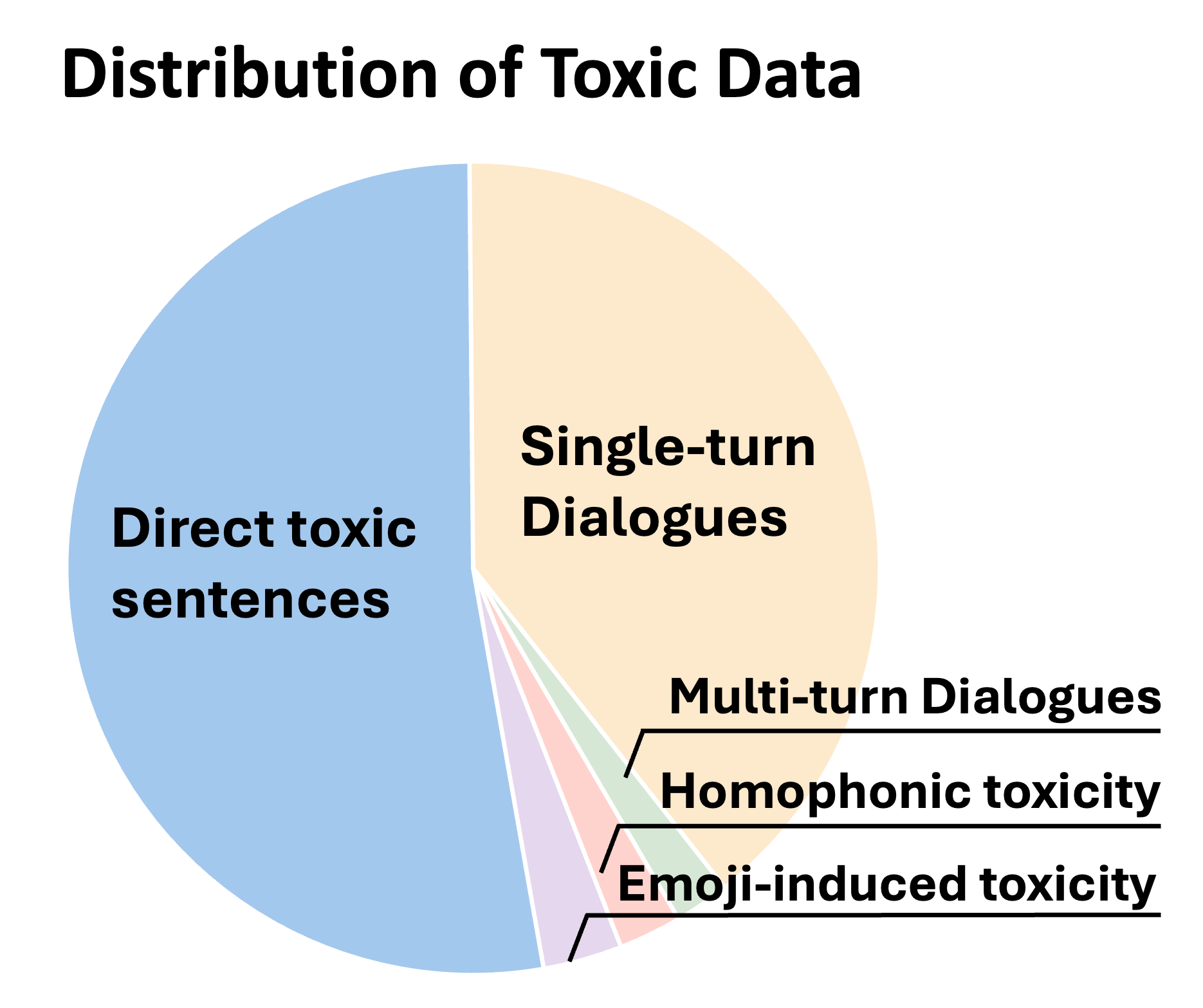}
  \caption{\textbf{Distribution of toxic data in the \textsc{ToxiRewriteCN} dataset}. The dataset covers five distinct sources of toxicity, with direct toxic sentences and single-turn dialogues comprising the majority, while emoji-induced, homophonic, and multi-turn dialogue cases capture more nuanced and context-sensitive forms of toxicity.}
  \label{fig:pie}
\end{figure}

\paragraph{Dataset Composition.}
The \textsc{ToxiRewriteCN} dataset covers a diverse range of toxicity scenarios across both sentence-level and conversation-level contexts. As shown in Figure~\ref{fig:pie}, over half of the dataset consists of \textbf{direct toxic sentences} (52.63\%), followed by \textbf{single-turn dialogues} (39.52\%). More nuanced forms of toxicity—such as \textbf{emoji-induced} (3.15\%), \textbf{homophonic toxicity} (2.51\%), and \textbf{multi-turn dialogues} (2.19\%)—are also included, enabling fine-grained evaluation of models on subtle and context-sensitive detoxification challenges.

\begin{figure}[t]
\centering
  \includegraphics[width=\columnwidth]{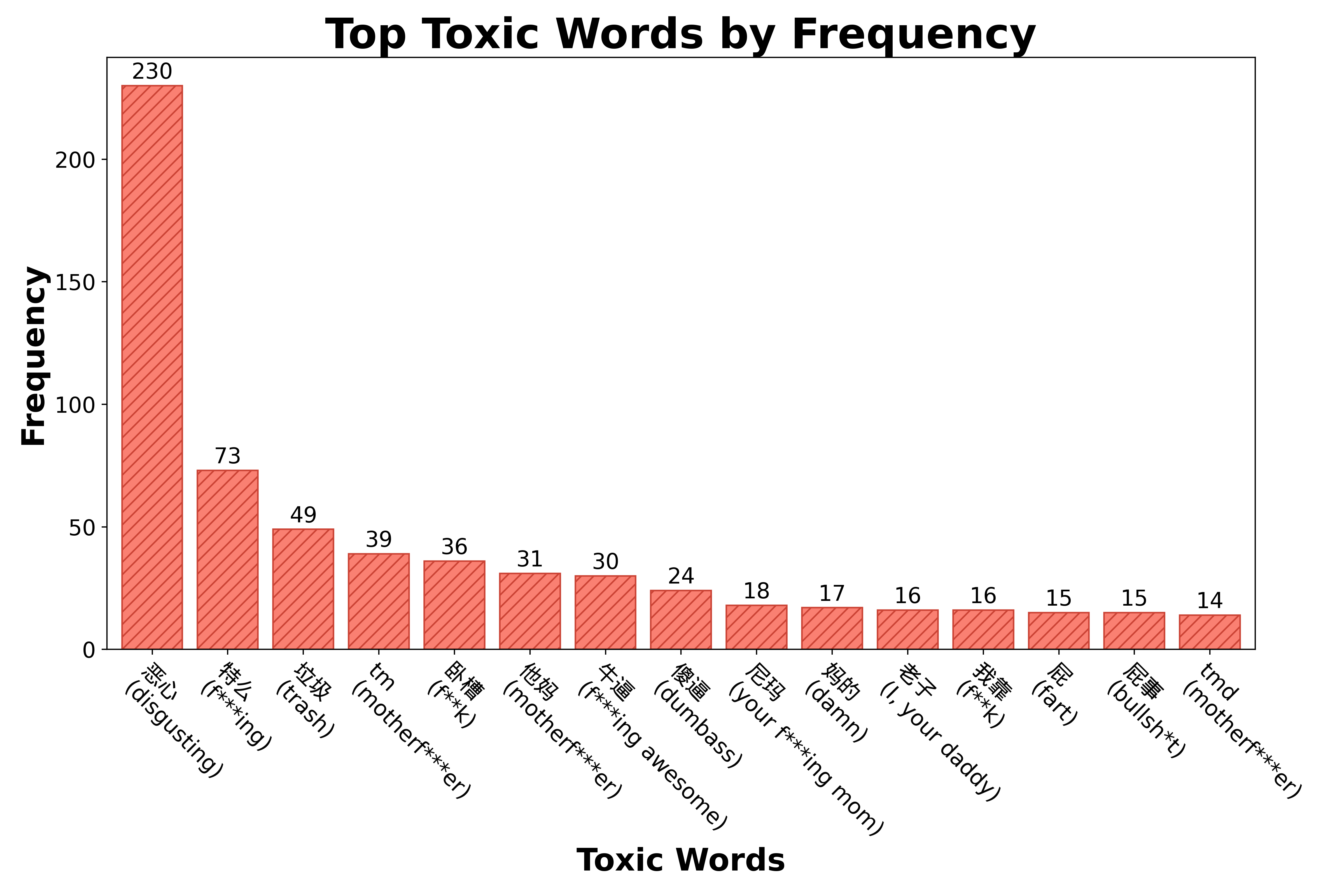}
  \caption{\textbf{Top 15 most frequent toxic words in the \textsc{ToxiRewriteCN} dataset}. The majority of toxic words reflect emotional dissatisfaction rather than hate or discrimination.}
  \label{fig:dis}
\end{figure}

We further analyze the distribution of toxic spans within the dataset (Figure ~\ref{fig:dis}). The most frequent toxic word is \begin{CJK}{UTF8}{gbsn}“恶心”\end{CJK} (“disgusting”), appearing 230 times. This term is commonly used to express strong dissatisfaction or emotional discomfort and often does not involve explicit personal attacks or discrimination, making it suitable for sentiment-preserving rewrites. Another frequent token is \begin{CJK}{UTF8}{gbsn}“卧槽”\end{CJK} (a colloquial expletive akin to “damn” or “WTF”), which appears 36 times. Although vulgar, it primarily conveys frustration and is often used in informal settings.

We also observe words such as\begin{CJK}{UTF8}{gbsn}“傻逼”\end{CJK} (“idiot” or stronger), which is semantically ambiguous. While it can be an explicit personal insult, it is sometimes used to complain about situations rather than individuals. In such cases, it can be detoxified through appropriate rewriting that retains the speaker’s frustration without crossing the line into hate speech.

\section{Performance Metrics of Different Scenarios}
\label{sec:diff_scenarios}

This section provides detailed evaluation scores for all models under different input perturbation categories, including single-sentence (Table \ref{tab:single-sentence}), emoji (Table \ref{tab:emoji}), homophone (Table \ref{tab:homophone}), single-turn conversation (Table \ref{tab:single-turn}) and multi-turn conversation (Table \ref{tab:multi-turn}) toxic rewrites. Metrics are fully reported across detoxification, fluency, content preservation, and sentiment polarity.

\begin{table*}[!t]
\centering
\scriptsize
\setlength\tabcolsep{4pt}
\renewcommand{\arraystretch}{1.2}
\begin{tabular}{l|ccc|cccc|c|ccc}
\toprule
\multirow{2}{*}{\textbf{Model}}
  & \multicolumn{3}{c|}{\textbf{Detox. Acc.}}
  & \multicolumn{4}{c|}{\textbf{Fluency}}
  & \multirow{2}{*}{\textbf{CntPres.$\uparrow$}}
  & \multicolumn{3}{c|}{\textbf{Sentiment Polarity}}
  \\ 
\cline{2-8}\cline{10-12}\addlinespace[0.5ex] 
  & \textbf{S-CLS$\uparrow$ } & \textbf{W-Clean$\uparrow$ } & \textbf{S-Clean$\uparrow$}
  & \textbf{BLEU$\uparrow$ } & \textbf{ChrF++$\uparrow$ } & \textbf{BS\_F1$\uparrow$ } & \textbf{COM.$\uparrow$}
  & 
  & \textbf{Toxic$\downarrow$ } & \textbf{Neutral$\uparrow$ } & \textbf{Polite $\downarrow$}\\ 
\midrule
\multicolumn{12}{@{}c@{}}{\textbf{Closed-Source Models}} \\ \midrule
\multicolumn{12}{@{}l@{}}{\textbf{Generation Models}} \\
\rowcolor{purple}
\hspace*{0.7em}|-- GPT-4o            & \closehigh{96.58} & 99.52& 99.39& 41.25& 31.46& 83.69& 86.1& 90.73& \closehigh{6.23} & \closehigh{79.00} & 14.77
 \\
\rowcolor{purple}
\hspace*{0.7em}|-- Qwen-Max          & 93.04& 97.81& 97.68& 48.14& 37.66& 86.06& 87.68& 91.44& 12.58& 77.29& 10.13
 \\
\rowcolor{purple}
\hspace*{0.7em}|-- Gemini-2.5-Flash  & 87.91& 96.57& 96.09& \closehigh{66.97} & \closehigh{53.91} & \closehigh{88.37} & \closehigh{88.28} & \closehigh{93.08} & 24.42& 69.35& \closehigh{6.23}
 \\
\rowcolor{purple}
\hspace*{0.7em}|-- Deepseek-V3       & 95.60 & \closehigh{99.62} & \closehigh{99.63} & 55.21& 43.87& 85.25& 86.67& 90.85& 8.30 & 77.29& 14.41 \\

\multicolumn{12}{@{}l@{}}{\textbf{Reasoning Models}} \\
\rowcolor{lighterpurple}
\hspace*{0.7em}|-- GPT-o1            & 90.48& 96.28& 95.97& 42.88& 31.88& 84.41& 87.12& 92.20 & 19.66& 74.11& \closehigh{6.23}
 \\
\rowcolor{lighterpurple}
\hspace*{0.7em}|-- Deepseek-R1       & 95.85& 98.28& 98.17& 34.19& 28.16& 80.12& 82.60 & 88.47& 9.40 & 65.32& 25.27
 \\
\rowcolor{lighterpurple}
\hspace*{0.7em}|-- Gemini-2.5-Pro    & 94.51& 98.57& 98.53& 43.30 & 36.84& 82.57& 84.91& 90.52& 13.92& 75.46& 10.62
 \\
\rowcolor{lighterpurple}
\hspace*{0.7em}|-- QwQ-32b           & 85.71& 97.04& 96.70 & 52.02& 39.16& 85.09& 86.51& 91.97& 26.25& 65.08& 8.67
 \\
\rowcolor{lighterpurple}
\hspace*{0.7em}|-- Qwen3-235B-A22B   & 91.21& 98.67& 98.41& 30.60 & 24.32& 79.74& 82.72& 88.90 & 17.70 & 64.59& 17.70  \\
\midrule
\multicolumn{12}{@{}c@{}}{\textbf{Open-Source Models}} \\ \midrule
\multicolumn{12}{@{}l@{}}{\textbf{MOE Models}} \\
\rowcolor{darkerblue}
\hspace*{0.7em}|-- Llama4 Maverick   & 83.27& 91.99& 90.84& 49.09& 37.03& 83.98& 85.48& 90.85& 25.52& 63.49& 10.99
 \\

\rowcolor{darkerblue}
\hspace*{0.7em}|-- Llama4 Scout      & 86.20 & 95.52& 94.99& 43.97& 32.21& 82.92& 84.93& 90.44& 25.40 & 66.42& 8.18
 \\

\rowcolor{darkerblue}
\hspace*{0.7em}|-- Qwen3-235B-A22B   & 89.26& 82.84& 81.20 & 63.32& 46.51& 85.28& 85.05& 91.08& 19.54& 69.11& 11.36
 \\

\rowcolor{darkerblue}
\hspace*{0.7em}|-- Qwen3-30B-A3B     & \openhigh{90.72} & 82.75& 81.32& \openhigh{64.26} & \openhigh{47.10} & 85.33& 84.99& 91.03& \openhigh{15.38} & \openhigh{72.04} & 12.58 \\

\multicolumn{12}{@{}l@{}}{\textbf{Dense Models}} \\
\rowcolor{lightblue}
\hspace*{0.7em}|-- Llama3-8B         & 76.07& \openhigh{97.62} & \openhigh{97.56} & 48.42& 35.95& 85.76 & 86.81& 91.42& 35.41& 52.99& 11.60  \\
\rowcolor{lightblue}
\hspace*{0.7em}|-- Llama3-3B         & 76.19& 96.85& 96.21& 49.08& 34.60 & 84.98& 86.24& 90.42& 34.07& 54.33& 11.60  \\
\rowcolor{lightblue}
\hspace*{0.7em}|-- Qwen3-8B          & 89.87& 94.85& 94.26& 41.84& 29.90 & 83.02& 85.41& 90.38& 15.51& 71.06& 13.43  \\
\rowcolor{lightblue}
\hspace*{0.7em}|-- Qwen3-4B          & 78.27& 84.46& 82.66& 60.47& 42.93& \openhigh{86.81} & \openhigh{86.91} & \openhigh{92.81} & 30.77& 61.29& \openhigh{7.94}  \\
\bottomrule
\end{tabular}
\caption{\textbf{Single-sentence performance metrics of various models across detoxification, fluency, content preservation, and sentiment polarity}. \closehigh{Box} highlights the best performance for each metric among closed-source models, while \openhigh{Box} highlights the best performance among open-source models.}
\label{tab:single-sentence}
\end{table*}

\begin{table*}[!b]
\centering
\scriptsize
\setlength\tabcolsep{4pt}
\renewcommand{\arraystretch}{1.2}
\begin{tabular}{l|ccc|cccc|c|ccc}
\toprule
\multirow{2}{*}{\textbf{Model}}
  & \multicolumn{3}{c|}{\textbf{Detox. Acc.}}
  & \multicolumn{4}{c|}{\textbf{Fluency}}
  & \multirow{2}{*}{\textbf{CntPres.$\uparrow$}}
  & \multicolumn{3}{c}{\textbf{Sentiment Polarity}} \\ 
\cline{2-8}\cline{10-12}\addlinespace[0.5ex]
  & \textbf{S-CLS$\uparrow$} & \textbf{W-Clean$\uparrow$} & \textbf{S-Clean$\uparrow$}
  & \textbf{BLEU$\uparrow$} & \textbf{ChrF++$\uparrow$} & \textbf{BS\_F1$\uparrow$} & \textbf{COM.$\uparrow$}
  &
  & \textbf{Toxic$\downarrow$} & \textbf{Neutral$\uparrow$} & \textbf{Polite$\downarrow$} \\
\midrule
\multicolumn{12}{@{}c@{}}{\textbf{Closed-Source Models}} \\ \midrule
\multicolumn{12}{@{}l@{}}{\textbf{Generation Models}} \\
\rowcolor{purple}
\hspace*{0.7em}|-- GPT-4o            & 75.51 & 95.31 & 93.88 & 47.55 & 39.09 & 82.71 & 82.75 & 89.20 & 40.82 & 51.02 & 8.16 \\
\rowcolor{purple}
\hspace*{0.7em}|-- Qwen-Max          & 59.18 & 92.19 & 89.80 & 58.01 & 52.23 & 85.05 & 84.51 & 90.10 & 53.06 & 38.78 & 8.16 \\
\rowcolor{purple}
\hspace*{0.7em}|-- Gemini-2.5-Flash  & 32.65 & 85.94 & 81.63 & \closehigh{69.79} & 52.35 & \closehigh{87.21} & \closehigh{85.94} & 91.62 & 73.47 & 26.53 & \closehigh{0.00} \\
\rowcolor{purple}
\hspace*{0.7em}|-- Deepseek-V3       & 75.51 & \closehigh{100.00} & \closehigh{100.00} & 56.12 & 44.73 & 85.34 & 85.69 & 88.97 & 34.69 & 55.10 & 10.20 \\

\multicolumn{12}{@{}l@{}}{\textbf{Reasoning Models}} \\
\rowcolor{lighterpurple}
\hspace*{0.7em}|-- GPT-o1            & 36.73 & 87.50 & 83.67 & 68.03 & \closehigh{54.39} & 87.03 & 85.36 & \closehigh{92.31} & 69.39 & 28.57 & 2.04 \\
\rowcolor{lighterpurple}
\hspace*{0.7em}|-- Deepseek-R1       & \closehigh{77.55} & 93.75 & 91.84 & 44.67 & 33.58 & 81.94 & 81.30 & 86.60 & \closehigh{32.65} & \closehigh{59.18} & 8.16 \\
\rowcolor{lighterpurple}
\hspace*{0.7em}|-- Gemini-2.5-Pro    & \closehigh{77.55} & 96.88 & 95.92 & 58.07 & 50.16 & 84.97 & 85.15 & 89.34 & 46.94 & 48.98 & 4.08 \\
\rowcolor{lighterpurple}
\hspace*{0.7em}|-- QwQ-32b           & 24.49 & 85.94 & 81.63 & 66.63 & 52.70 & 85.90 & 83.70 & 91.97 & 81.63 & 16.33 & 2.04 \\
\rowcolor{lighterpurple}
\hspace*{0.7em}|-- Qwen3-235B-A22B   & 53.06 & 95.31 & 93.88 & 62.04 & 46.20 & 86.02 & 83.73 & 91.56 & 57.14 & 40.82 & 2.04 \\

\midrule
\multicolumn{12}{@{}c@{}}{\textbf{Open-Source Models}} \\ \midrule
\multicolumn{12}{@{}l@{}}{\textbf{MOE Models}} \\
\rowcolor{darkerblue}
\hspace*{0.7em}|-- Llama4 Maverick   & 34.69 & 85.94 & 81.63 & 58.24 & 39.35 & 83.85 & 83.56 & 89.84 & 69.39 & 30.61 & \openhigh{0.00} \\
\rowcolor{darkerblue}
\hspace*{0.7em}|-- Llama4 Scout      & 36.73 & \openhigh{90.62} & \openhigh{87.76} & 55.99 & 38.28 & 85.00 & 83.94 & 90.91 & 67.35 & 30.61 & 2.04 \\
\rowcolor{darkerblue}
\hspace*{0.7em}|-- Qwen3-235B-A22B   & 34.69 & \openhigh{90.62} & \openhigh{87.76} & \openhigh{64.38} & \openhigh{50.82} & \openhigh{86.51} & \openhigh{84.03} & \openhigh{91.36} & 71.43 & 26.53 & 2.04 \\
\rowcolor{darkerblue}
\hspace*{0.7em}|-- Qwen3-30B-A3B     & 44.90 & 89.06 & 85.71 & 64.19 & 47.37 & 85.32 & 82.51 & 90.27 & 55.10 & 38.78 & 6.12 \\

\multicolumn{12}{@{}l@{}}{\textbf{Dense Models}} \\
\rowcolor{lightblue}
\hspace*{0.7em}|-- Llama3-8B         & 40.82 & 89.06 & 85.71 & 58.75 & 47.12 & 82.87 & 79.59 & 88.38 & 67.35 & 26.53 & 6.12 \\
\rowcolor{lightblue}
\hspace*{0.7em}|-- Llama3-3B         & 40.82 & 89.06 & 85.71 & 58.09 & 46.27 & 82.69 & 79.89 & 88.23 & 67.35 & 26.53 & 6.12 \\
\rowcolor{lightblue}
\hspace*{0.7em}|-- Qwen3-8B          & \openhigh{57.14} & \openhigh{90.62} & \openhigh{87.76} & 45.32 & 34.57 & 81.51 & 82.73 & 89.22 & \openhigh{44.90} & \openhigh{46.94} & 8.16 \\
\rowcolor{lightblue}
\hspace*{0.7em}|-- Qwen3-4B          & 59.18 & 89.06 & 85.71 & 59.33 & 46.41 & 84.25 & 82.82 & 91.01 & 51.02 & 36.73 & 12.24 \\
\bottomrule
\end{tabular}
\caption{\textbf{Emoji performance metrics of various models across detoxification, fluency, content preservation, and sentiment polarity.} \closehigh{Box} highlights the best performance for each metric among closed-source models, while \openhigh{Box} highlights the best performance among open-source models.}
\label{tab:emoji}
\end{table*}

\begin{table*}[!t]
\centering
\scriptsize
\setlength\tabcolsep{4pt}
\renewcommand{\arraystretch}{1.2}
\begin{tabular}{l|ccc|cccc|c|ccc}
\toprule
\multirow{2}{*}{\textbf{Model}}
  & \multicolumn{3}{c|}{\textbf{Detox. Acc.}}
  & \multicolumn{4}{c|}{\textbf{Fluency}}
  & \multirow{2}{*}{\textbf{CntPres.$\uparrow$}}
  & \multicolumn{3}{c}{\textbf{Sentiment Polarity}} \\ 
\cline{2-8}\cline{10-12}\addlinespace[0.5ex]
  & \textbf{S-CLS$\uparrow$} & \textbf{W-Clean$\uparrow$} & \textbf{S-Clean$\uparrow$}
  & \textbf{BLEU$\uparrow$} & \textbf{ChrF++$\uparrow$} & \textbf{BS\_F1$\uparrow$} & \textbf{COM.$\uparrow$}
  &
  & \textbf{Toxic$\downarrow$} & \textbf{Neutral$\uparrow$} & \textbf{Polite$\downarrow$} \\
\midrule
\multicolumn{12}{@{}c@{}}{\textbf{Closed-Source Models}} \\ \midrule
\multicolumn{12}{@{}l@{}}{\textbf{Generation Models}} \\
\rowcolor{purple}
\hspace*{0.7em}|-- GPT-4o            & \closehigh{82.05} & \closehigh{98.15} & \closehigh{97.44} & 45.42 & 36.28 & 81.76 & 82.91 & 86.14 & 41.03 & 41.03 & 17.95 \\
\rowcolor{purple}
\hspace*{0.7em}|-- Qwen-Max          & 69.23 & 94.44 & 92.31 & 52.62 & 47.48 & 83.89 & 83.10 & 88.76 & \closehigh{30.77} & \closehigh{61.54} & 7.69 \\
\rowcolor{purple}
\hspace*{0.7em}|-- Gemini-2.5-Flash  & 56.41 & 88.89 & 84.62 & \closehigh{66.07} & \closehigh{49.40} & \closehigh{85.65} & 84.22 & \closehigh{90.13} & 58.97 & 41.03 & \closehigh{0.00} \\
\rowcolor{purple}
\hspace*{0.7em}|-- Deepseek-V3       & 79.49 & \closehigh{98.15} & \closehigh{97.44} & 57.70 & 44.93 & 84.40 & \closehigh{84.88} & 88.14 & 41.03 & 48.72 & 10.26 \\

\multicolumn{12}{@{}l@{}}{\textbf{Reasoning Models}} \\
\rowcolor{lighterpurple}
\hspace*{0.7em}|-- GPT-o1            & 46.15 & 88.89 & 84.62 & 61.75 & 48.20 & 85.26 & 84.29 & 89.65 & 66.67 & 30.77 & 2.56 \\
\rowcolor{lighterpurple}
\hspace*{0.7em}|-- Deepseek-R1       & 74.36 & 94.44 & 92.31 & 52.06 & 37.80 & 82.99 & 81.53 & 86.73 & 38.46 & 53.85 & 7.69 \\
\rowcolor{lighterpurple}
\hspace*{0.7em}|-- Gemini-2.5-Pro    & 66.67 & 96.30 & 94.87 & 59.21 & 47.17 & 84.53 & 84.40 & 89.25 & 43.59 & 53.85 & 2.56 \\
\rowcolor{lighterpurple}
\hspace*{0.7em}|-- QwQ-32b           & 41.03 & 90.74 & 87.18 & 56.96 & 45.66 & 83.72 & 80.72 & 88.68 & 71.79 & 25.64 & 2.56 \\
\rowcolor{lighterpurple}
\hspace*{0.7em}|-- Qwen3-235B-A22B   & 58.97 & 94.44 & 92.31 & 60.87 & 44.50 & 84.09 & 81.88 & 88.56 & 46.15 & 48.72 & 5.13 \\

\midrule
\multicolumn{12}{@{}c@{}}{\textbf{Open-Source Models}} \\ \midrule
\multicolumn{12}{@{}l@{}}{\textbf{MOE Models}} \\
\rowcolor{darkerblue}
\hspace*{0.7em}|-- Llama4 Maverick   & 46.15 & 85.19 & 79.49 & 56.27 & 37.80 & 82.30 & 81.38 & 88.40 & 61.54 & 33.33 & 5.13 \\
\rowcolor{darkerblue}
\hspace*{0.7em}|-- Llama4 Scout      & 35.90 & 88.89 & 84.62 & 52.43 & 39.82 & 83.52 & 81.59 & 89.34 & 76.92 & 23.08 & \openhigh{0.00} \\
\rowcolor{darkerblue}
\hspace*{0.7em}|-- Qwen3-235B-A22B   & 48.72 & \openhigh{92.59} & \openhigh{89.74} & 56.11 & 41.60 & 83.78 & 81.84 & 88.82 & 64.10 & 25.64 & 10.26 \\
\rowcolor{darkerblue}
\hspace*{0.7em}|-- Qwen3-30B-A3B     & 51.28 & 85.19 & 79.49 & \openhigh{63.80} & \openhigh{47.49} & \openhigh{85.20} & \openhigh{82.79} & 90.14 & 56.41 & 33.33 & 10.26 \\

\multicolumn{12}{@{}l@{}}{\textbf{Dense Models}} \\
\rowcolor{lightblue}
\hspace*{0.7em}|-- Llama3-8B         & 58.97 & 81.48 & 76.92 & 48.82 & 33.90 & 78.63 & 74.07 & 86.18 & 46.15 & 38.46 & 15.38 \\
\rowcolor{lightblue}
\hspace*{0.7em}|-- Llama3-3B         & 56.41 & 81.48 & 76.92 & 51.25 & 40.68 & 80.76 & 76.74 & 87.50 & 48.72 & \openhigh{41.03} & 10.26 \\
\rowcolor{lightblue}
\hspace*{0.7em}|-- Qwen3-8B          & \openhigh{69.23} & 85.19 & 82.05 & 45.72 & 33.92 & 81.62 & 81.20 & 89.78 & \openhigh{41.03} & \openhigh{41.03} & 17.95 \\
\rowcolor{lightblue}
\hspace*{0.7em}|-- Qwen3-4B          & 58.97 & 75.93 & 69.23 & 56.23 & 43.37 & 83.59 & 82.26 & \openhigh{91.69} & 53.85 & 33.33 & 12.82 \\
\bottomrule
\end{tabular}
\caption{\textbf{Homophone performance metrics of various models across detoxification, fluency, content preservation, and sentiment polarity.} \closehigh{Box} highlights the best performance for each metric among closed-source models, while \openhigh{Box} highlights the best performance among open-source models.}
\label{tab:homophone}

\end{table*}

\begin{table*}[!b]
\centering
\scriptsize
\setlength\tabcolsep{4pt}
\renewcommand{\arraystretch}{1.2}
\begin{tabular}{l|ccc|cccc|c|ccc}
\toprule
\multirow{2}{*}{\textbf{Model}}
  & \multicolumn{3}{c|}{\textbf{Detox. Acc.}}
  & \multicolumn{4}{c|}{\textbf{Fluency}}
  & \multirow{2}{*}{\textbf{CntPres.$\uparrow$}}
  & \multicolumn{3}{c}{\textbf{Sentiment Polarity}} \\ 
\cline{2-8}\cline{10-12}\addlinespace[0.5ex]
  & \textbf{S-CLS$\uparrow$ } & \textbf{W-Clean$\uparrow$ } & \textbf{S-Clean$\uparrow$}
  & \textbf{BLEU$\uparrow$ } & \textbf{ChrF++$\uparrow$ } & \textbf{BS\_F1$\uparrow$ } & \textbf{COM.$\uparrow$}
  &
  & \textbf{Toxic$\downarrow$ } & \textbf{Neutral$\uparrow$ } & \textbf{Polite $\downarrow$} \\
\midrule
\multicolumn{12}{@{}c@{}}{\textbf{Closed-Source Models}} \\ \midrule
\multicolumn{12}{@{}l@{}}{\textbf{Generation Models}} \\
\rowcolor{purple}
\hspace*{0.7em}|-- GPT-4o            & \closehigh{80.49} & 94.45 & 96.00 & 84.54 & 75.23 & 94.56 & 90.38 & 98.56 & \closehigh{28.78} & \closehigh{55.93} & 15.28 \\
\rowcolor{purple}
\hspace*{0.7em}|-- Qwen-Max          & 76.59 & 91.85 & 94.16 & 88.47 & 80.01 & 95.76 & 91.21 & 98.90 & 36.59 & 51.38 & 12.03 \\
\rowcolor{purple}
\hspace*{0.7em}|-- Gemini-2.5-Flash  & 64.88 & 83.19 & 87.19 & \closehigh{93.12} & 82.77 & \closehigh{96.98} & \closehigh{91.45} & \closehigh{99.37} & 46.50 & 49.27 & 4.23 \\
\rowcolor{purple}
\hspace*{0.7em}|-- Deepseek-V3       & 74.80 & 90.12 & 93.14 & 91.86 & \closehigh{82.79} & 96.31 & 91.34 & 98.95 & 34.15 & 50.08 & 15.77 \\

\multicolumn{12}{@{}l@{}}{\textbf{Reasoning Models}} \\
\rowcolor{lighterpurple}
\hspace*{0.7em}|-- GPT-o1            & 55.12 & 92.37 & 93.48 & 90.78 & 81.40 & 96.67 & 91.44 & 99.32 & 56.91 & 39.67 & \closehigh{3.41} \\
\rowcolor{lighterpurple}
\hspace*{0.7em}|-- Deepseek-R1       & 75.93 & 96.88 & 97.60 & 82.45 & 72.84 & 94.09 & 89.92 & 98.31 & 33.66 & 48.13 & 18.21 \\
\rowcolor{lighterpurple}
\hspace*{0.7em}|-- Gemini-2.5-Pro    & 65.20 & \closehigh{98.09} & \closehigh{98.17} & 89.18 & 80.31 & 95.93 & 90.92 & 99.08 & 48.78 & 47.15 & 4.07 \\
\rowcolor{lighterpurple}
\hspace*{0.7em}|-- QwQ-32b           & 65.53 & 91.33 & 93.94 & 88.49 & 77.49 & 95.88 & 90.89 & 99.01 & 45.69 & 46.99 & 7.32 \\
\rowcolor{lighterpurple}
\hspace*{0.7em}|-- Qwen3-235B-A22B   & 73.17 & 93.07 & 94.74 & 86.26 & 76.80 & 94.87 & 90.31 & 98.62 & 36.42 & 51.38 & 12.20 \\

\midrule
\multicolumn{12}{@{}c@{}}{\textbf{Open-Source Models}} \\ \midrule
\multicolumn{12}{@{}l@{}}{\textbf{MOE Models}} \\
\rowcolor{darkerblue}
\hspace*{0.7em}|-- Llama4 Maverick   & 72.52 & 85.58 & 84.88 & 88.13 & 77.09 & 94.99 & 90.37 & 98.59 & 35.28 & 41.95 & 22.76 \\
\rowcolor{darkerblue}
\hspace*{0.7em}|-- Llama4 Scout      & 66.50 & 90.73 & 88.78 & 75.18 & 63.93 & 90.13 & 88.16 & 97.42 & 46.83 & 43.58 & 9.59 \\
\rowcolor{darkerblue}
\hspace*{0.7em}|-- Qwen3-235B-A22B   & 69.92 & 83.75 & 83.74 & 80.53 & 70.86 & 89.89 & 85.57 & 95.81 & 42.60 & \openhigh{44.88} & 12.52 \\
\rowcolor{darkerblue}
\hspace*{0.7em}|-- Qwen3-30B-A3B     & 66.99 & 83.98 & 83.58 & 80.70 & 70.92 & 90.02 & 85.73 & 95.94 & 42.11 & 42.76 & 15.12 \\

\multicolumn{12}{@{}l@{}}{\textbf{Dense Models}} \\
\rowcolor{lightblue}
\hspace*{0.7em}|-- Llama3-8B         & \openhigh{75.77} & \openhigh{91.99} & \openhigh{92.20} & 89.24 & 78.67 & 95.33 & 90.66 & 98.78 & \openhigh{30.24} & 34.31 & 35.45 \\
\rowcolor{lightblue}
\hspace*{0.7em}|-- Llama3-3B         & 75.28 & 82.61 & 82.11 & 90.17 & 80.66 & 95.59 & 90.57 & 98.70 & 30.89 & 33.98 & 35.12 \\
\rowcolor{lightblue}
\hspace*{0.7em}|-- Qwen3-8B          & 57.40 & 71.17 & 72.52 & \openhigh{91.71} & \openhigh{81.06} & \openhigh{96.39} & \openhigh{90.92} & \openhigh{99.15} & 52.52 & 38.21 & \openhigh{9.27} \\
\rowcolor{lightblue}
\hspace*{0.7em}|-- Qwen3-4B          & 58.54 & 62.01 & 65.20 & 89.05 & 78.03 & 94.66 & 89.90 & 98.82 & 48.78 & 35.45 & 15.77 \\
\bottomrule
\end{tabular}
\caption{\textbf{Single-turn conversation performance metrics of various models across detoxification, fluency, content preservation, and sentiment polarity.} \closehigh{Box} highlights the best performance for each metric among closed-source models, while \openhigh{Box} highlights the best performance among open-source models.}
\label{tab:single-turn}
\end{table*}

\begin{table*}[t]
\centering
\scriptsize
\setlength\tabcolsep{4pt}
\renewcommand{\arraystretch}{1.2}
\begin{tabular}{l|ccc|cccc|c|ccc}
\toprule
\multirow{2}{*}{\textbf{Model}}
  & \multicolumn{3}{c|}{\textbf{Detox. Acc.}}
  & \multicolumn{4}{c|}{\textbf{Fluency}}
  & \multirow{2}{*}{\textbf{CntPres.$\uparrow$}}
  & \multicolumn{3}{c}{\textbf{Sentiment Polarity}} \\ 
\cline{2-8}\cline{10-12}\addlinespace[0.5ex]
  & \textbf{S-CLS$\uparrow$ } & \textbf{W-Clean$\uparrow$ } & \textbf{S-Clean$\uparrow$}
  & \textbf{BLEU$\uparrow$ } & \textbf{ChrF++$\uparrow$ } & \textbf{BS\_F1$\uparrow$ } & \textbf{COM.$\uparrow$}
  &
  & \textbf{Toxic$\downarrow$ } & \textbf{Neutral$\uparrow$ } & \textbf{Polite $\downarrow$} \\
\midrule
\multicolumn{12}{@{}c@{}}{\textbf{Closed-Source Models}} \\ \midrule
\multicolumn{12}{@{}l@{}}{\textbf{Generation Models}} \\
\rowcolor{purple}
\hspace*{0.7em}|-- GPT-4o            & 52.94 & 87.50 & 73.53 & 77.90 & 70.65 & 92.94 & 87.30 & 98.90 & 58.82 & 38.24 & 2.94 \\
\rowcolor{purple}
\hspace*{0.7em}|-- Qwen-Max          & 55.88 & 86.25 & 70.59 & 88.61 & 80.67 & 95.94 & \closehigh{88.86} & 99.10 & 64.71 & 32.35 & 2.94 \\
\rowcolor{purple}
\hspace*{0.7em}|-- Gemini-2.5-Flash  & 47.06 & 86.25 & 76.47 & 90.16 & 79.82 & 95.78 & 88.23 & 99.06 & 64.71 & 32.35 & 2.94 \\
\rowcolor{purple}
\hspace*{0.7em}|-- Deepseek-V3       & 64.71 & 82.50 & 73.53 & \closehigh{90.38} & \closehigh{83.46} & \closehigh{96.10} & 88.55 & 99.16 & 55.88 & 38.24 & 5.88 \\

\multicolumn{12}{@{}l@{}}{\textbf{Reasoning Models}} \\
\rowcolor{lighterpurple}
\hspace*{0.7em}|-- GPT-o1            & 52.94 & 91.25 & 79.41 & 87.32 & 81.14 & 95.41 & 88.55 & \closehigh{99.18} & 82.35 & 17.65 & \closehigh{0.00} \\
\rowcolor{lighterpurple}
\hspace*{0.7em}|-- Deepseek-R1       & \closehigh{76.47} & 92.50 & 85.29 & 78.60 & 73.32 & 91.85 & 85.93 & 98.21 & \closehigh{29.41} & \closehigh{55.88} & 14.71 \\
\rowcolor{lighterpurple}
\hspace*{0.7em}|-- Gemini-2.5-Pro    & 55.88 & \closehigh{98.75} & \closehigh{97.06} & 80.67 & 74.39 & 93.02 & 87.49 & 98.74 & 64.71 & 32.35 & 2.94 \\
\rowcolor{lighterpurple}
\hspace*{0.7em}|-- QwQ-32b           & 64.71 & 93.75 & 85.29 & 84.93 & 76.23 & 93.60 & 87.37 & 98.85 & 50.00 & 38.24 & 11.76 \\
\rowcolor{lighterpurple}
\hspace*{0.7em}|-- Qwen3-235B-A22B   & 61.76 & 91.25 & 85.29 & 79.62 & 71.16 & 92.79 & 86.96 & 98.72 & 47.06 & 44.12 & 8.82 \\

\midrule
\multicolumn{12}{@{}c@{}}{\textbf{Open-Source Models}} \\ \midrule
\multicolumn{12}{@{}l@{}}{\textbf{MOE Models}} \\
\rowcolor{darkerblue}
\hspace*{0.7em}|-- Llama4 Maverick   & 41.18 & 72.50 & 58.82 & 77.85 & 67.26 & 92.55 & 87.45 & 98.55 & 67.65 & 26.47 & 5.88 \\
\rowcolor{darkerblue}
\hspace*{0.7em}|-- Llama4 Scout      & 50.00 & \openhigh{85.00} & 70.59 & 76.89 & 67.48 & 93.03 & 86.87 & 99.11 & 73.53 & 20.59 & 5.88 \\
\rowcolor{darkerblue}
\hspace*{0.7em}|-- Qwen3-235B-A22B   & 61.76 & 81.25 & 70.59 & 82.37 & 75.16 & 94.27 & 87.08 & 99.01 & 61.76 & 29.41 & 8.82 \\
\rowcolor{darkerblue}
\hspace*{0.7em}|-- Qwen3-30B-A3B     & 41.18 & 67.50 & 52.94 & 90.56 & 83.14 & 96.02 & 88.18 & 99.32 & 67.65 & \openhigh{32.35} & \openhigh{0.00} \\

\multicolumn{12}{@{}l@{}}{\textbf{Dense Models}} \\
\rowcolor{lightblue}
\hspace*{0.7em}|-- Llama3-8B         & 61.76 & 82.50 & 70.59 & 58.59 & 60.22 & 79.14 & 70.28 & 84.29 & 52.94 & 8.82 & 38.24 \\
\rowcolor{lightblue}
\hspace*{0.7em}|-- Llama3-3B         & \openhigh{64.71} & \openhigh{85.00} & \openhigh{73.53} & 48.87 & 54.63 & 74.95 & 65.69 & 80.75 & \openhigh{47.06} & 5.88 & 47.06 \\
\rowcolor{lightblue}
\hspace*{0.7em}|-- Qwen3-8B          & 41.18 & 61.25 & 38.24 & \openhigh{92.23} & \openhigh{86.01} & \openhigh{96.67} & \openhigh{88.52} & \openhigh{99.64} & 79.41 & 17.65 & 2.94 \\
\rowcolor{lightblue}
\hspace*{0.7em}|-- Qwen3-4B          & 32.35 & 38.75 & 20.59 & 91.64 & 84.36 & 96.32 & 88.06 & 99.60 & 82.35 & 14.71 & 2.94 \\
\bottomrule
\end{tabular}
\caption{\textbf{Multi-turn conversation performance metrics of various models across detoxification, fluency, content preservation, and sentiment polarity.} \closehigh{Box} highlights the best performance for each metric among closed-source models, while \openhigh{Box} highlights the best performance among open-source models.}
\label{tab:multi-turn}
\end{table*}

\section{Human Preference Analysis}
\label{sec:human}

\begin{figure}[t]
\centering
  \includegraphics[width=\columnwidth]{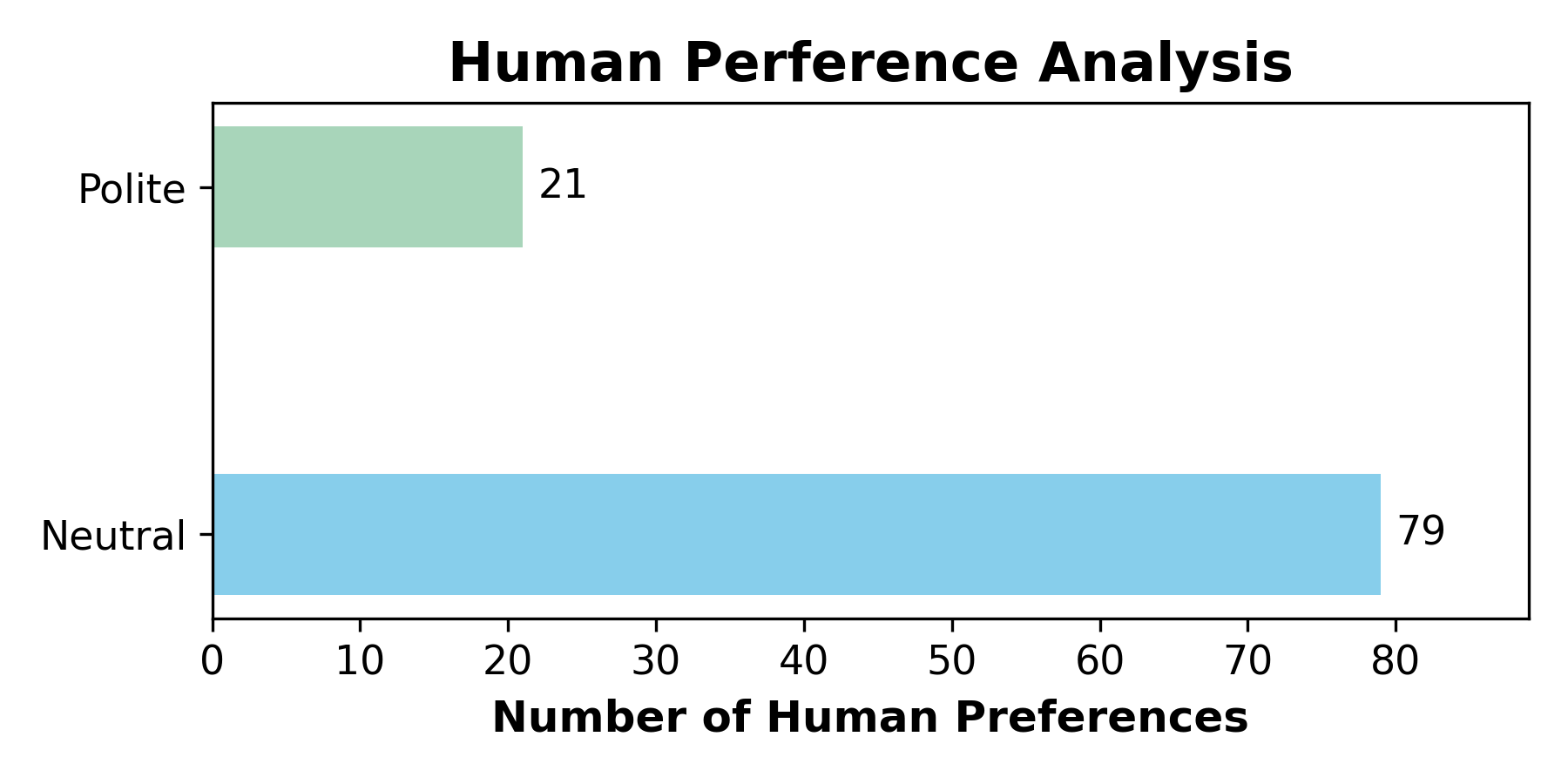}
  \caption{\textbf{Human preference comparison between sentiment-preserving (neutral) and over-polite detoxification rewrites.} Annotators significantly preferred neutral rewrites (79\%) over polite ones (21\%).}
  \label{fig:human}
\end{figure}

To assess whether sentiment-preserving rewrites are more aligned with human expectations, we conducted a preference study over 100 sampled examples spanning all five categories: sentence-level (52), emoji (3), homophone (3), single-turn (40), and multi-turn (2). For each example, annotators were shown two rewrites of the same toxic input: one neutral (preserving sentiment) and one polite (over-sanitized), generated by two competitive models—GPT-4o and Deepseek-V3—randomized in order. Annotators were asked to select the version that better preserved the original user intent while removing toxicity.

As shown in Figure ~\ref{fig:human}, out of 100 comparisons, the sentiment-preserving (neutral) rewrite was preferred in 79 cases, while the polite rewrite was chosen in 21 cases. This indicates a strong human preference for rewrites that retain emotional tone rather than overly formal rephrasings, reinforcing the core motivation of \textsc{ToxiRewriteCN}.

\section{Implementation Details of Classifiers}
\label{sec: impl_class}
We fine-tuned two classifiers based on the Qwen3-32B model: a toxicity classifier and a sentiment polarity classifier.
The toxicity classifier is a binary classifier that determines whether a rewritten sentence is toxic or non-toxic. We constructed the training dataset by combining toxic and non-toxic samples from the \textsc{ToxiRewriteCN} dataset with additional samples from the ToxiCN dataset, resulting in a total of 4,112 non-toxic and 4,035 toxic sentences. Training was performed using LoRA-based efficient fine-tuning, with the following hyperparameters:a LoRA rank of 8, LoRA alpha of 16, and a dropout rate of 0.05. The model was trained for 3 epochs using a learning rate of 2e-5 with a cosine learning rate scheduler.

The sentiment classifier is a three-class classifier that predicts whether a rewritten sentence conveys a toxic, neutral, or polite sentiment. The training data was constructed entirely from the \textsc{ToxiRewriteCN} dataset. To improve model performance and training stability, we adjusted the label distribution to a 1:2:1 ratio of toxic, neutral, and polite examples by duplicating neutral samples, resulting in a total of 6,224 training instances. The classifier was also fine-tuned using LoRA, with the same hyperparameter settings:a LoRA rank of 8, LoRA alpha of 16, and a dropout rate of 0.05. It was trained for 5 epochs with a learning rate of 2e-5 using a cosine scheduler. All training was conducted using 8 NVIDIA H100 GPUs.

\end{document}